\documentclass[10pt,twocolumn,letterpaper]{article}

\usepackage{graphicx}
\usepackage{subcaption}
\usepackage{float}
\usepackage[justification=raggedright]{caption}	%
\usepackage{lscape}                                         %

\usepackage[lined,ruled,linesnumbered]{algorithm2e}

\usepackage{booktabs}                   %
\usepackage{multirow}

\usepackage{paralist}
\usepackage{enumitem}

\usepackage{bm}                          %
\usepackage{epsfig}                      %
\usepackage{graphicx}                  %
\usepackage{times}
\usepackage{mathptmx}
\usepackage{mathtools}
\usepackage{amssymb,amsmath}   %

\usepackage{units}
\usepackage{color}

\usepackage{comment}

\usepackage{url}  %
\usepackage[pagebackref=true,breaklinks=true,letterpaper=true,colorlinks,bookmarks=false]{hyperref}

\usepackage{xspace}
\usepackage[table]{xcolor}
\usepackage{setspace}

\def\etal{et~al.\_}			  %

\newlength\paramargin
\newlength\figmargin
\newlength\secmargin
\newlength\figcapmargin
\newlength\tabmargin

\newcommand{\mpage}[2]
{
\begin{minipage}{#1\linewidth}\centering
#2
\end{minipage}
}

\newcommand{\figref}[1]{Figure~\ref{fig:#1}}

\long\def\ignorethis#1{}

\newcommand{\mfigure}[2]
{
\begin{minipage}{#1\linewidth}\centering
\includegraphics[width=\linewidth]{#2}
\end{minipage}
}

\def\xi{\mathbf{x}_i}

\graphicspath{{figure}, {example}}

\usepackage{iccv}

\iccvfinalcopy %

\def\iccvPaperID{409} %

\ificcvfinal\fi

\begin{document}

\title{Guided Image-to-Image Translation with Bi-Directional Feature Transformation}

\author{Badour AlBahar \qquad Jia-Bin Huang\\
Virginia Tech\\
\url{https://github.com/vt-vl-lab/Guided-pix2pix}
}

\twocolumn[{
\renewcommand\twocolumn[1][]{#1}
\maketitle
\begin{center}
\centering
\mpage{0.01}{\raisebox{0pt}{\rotatebox{90}{Pose Transfer}}} 
\mfigure{0.15}{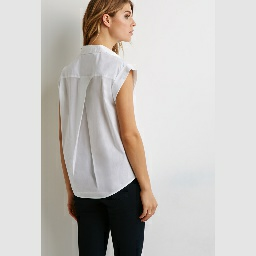} \hfill
\mfigure{0.15}{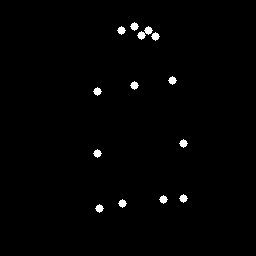} \hfill
\mfigure{0.15}{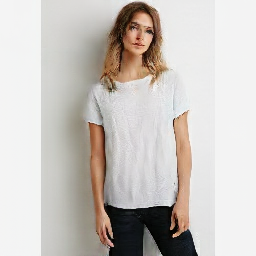} \hfill
\mfigure{0.15}{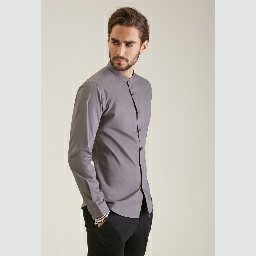} \hfill
\mfigure{0.15}{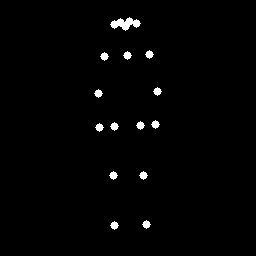} \hfill
\mfigure{0.15}{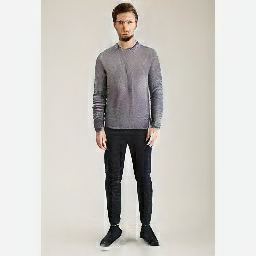} 

\mpage{0.01}{\raisebox{0pt}{\rotatebox{90}{Texture Transfer}}} 
\mfigure{0.15}{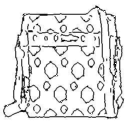} \hfill
\mfigure{0.15}{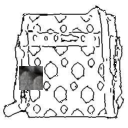} \hfill
\mfigure{0.15}{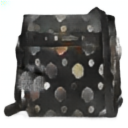} \hfill
\mfigure{0.15}{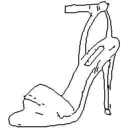} \hfill
\mfigure{0.15}{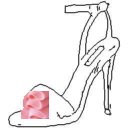} \hfill
\mfigure{0.15}{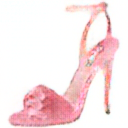}

\mpage{0.01}{\raisebox{0pt}{\rotatebox{90}{Upsampling}}} 
\mfigure{0.15}{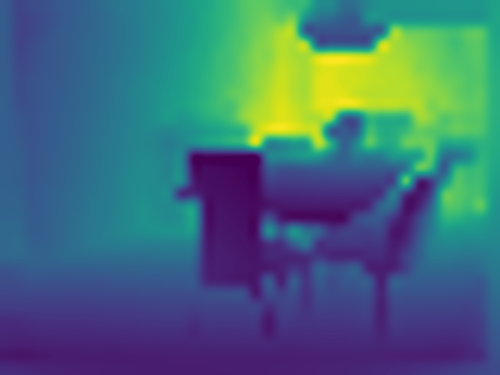} \hfill
\mfigure{0.15}{} \hfill
\mfigure{0.15}{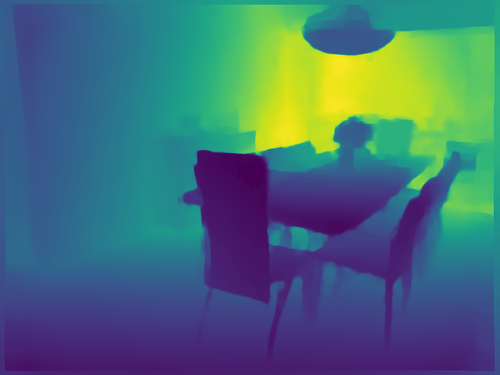} \hfill
\mfigure{0.15}{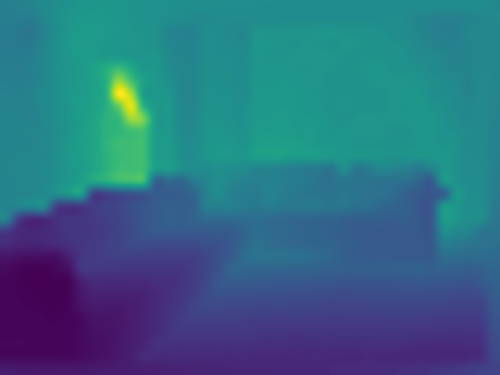} \hfill
\mfigure{0.15}{} \hfill
\mfigure{0.15}{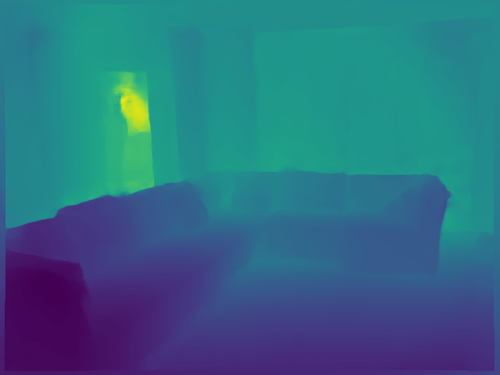}

~~~
\mpage{0.15}{Input} \hfill
\mpage{0.15}{Guide} \hfill
\mpage{0.15}{Ours} \hfill
\mpage{0.15}{Input} \hfill
\mpage{0.15}{Guide} \hfill
\mpage{0.15}{Ours} \\
\vspace{0.5mm}
\captionof{figure}{
\textbf{Applications of guided image-to-image translation.}
We present an algorithm that translates an input image into a corresponding output image while respecting the constraints specified in the provided guidance image.
These controllable image-to-image translation problems often require task-specific architectures and training objective functions as the guidance can take various different forms (e.g., color strokes, sketch, texture patch, image, and mask).
We introduce a new conditioning scheme for controlling image synthesis using available guidance signals and demonstrate applicability to several sample applications, including person image synthesis guided by a given pose (\emph{top}), sketch-to-photo synthesis guided with a texture patch (\emph{middle}), and depth upsampling guided with an RGB image (\emph{bottom}).
}
\label{fig:teaser}
\end{center}

}]
\ificcvfinal\thispagestyle{empty}\fi

\begin{abstract}
\vspace{-3mm}
We address the problem of guided image-to-image translation where we translate an input image into another while respecting the constraints provided by an external, user-provided guidance image.
Various conditioning methods for leveraging the given guidance image have been explored, including input/feature concatenation and conditional affine transformation of feature activations.
All these conditioning mechanisms, however, are uni-directional, i.e., no information flow from the input image back to the guidance.
To better utilize the constraints of the guidance image, we present a bi-directional feature transformation (bFT) scheme.
We show that our bFT scheme outperforms other conditioning schemes and has comparable results to state-of-the-art methods on different tasks.
\end{abstract}

\vspace{\secmargin}
\section{Introduction}
\label{sec:intro}

In an image-to-image translation problem~\cite{isola2017image}, we aim to translate an image from one domain to another. Many problems in computer vision, graphics, and image processing can be formulated as image-to-image translation tasks, including semantic image synthesis, style transfer, colorization, sketch to photos, to name a few. 
An extension to these image-to-image translation problems involves an additional \emph{guidance image} that helps achieve controllable translation. 
A guidance image typically reflects the desired visual effects or constraints specified by a user or provides additional information via other modalities (color/depth, flash/non-flash, color/IR). A guidance image can thus take many different forms, e.g. color strokes or palette, semantic labels, texture patch, image, or mask. As such, most of the existing solutions for such problems often have application-specific architectures and objective functions, and consequently cannot be directly applied to other problems.

The main technical question for guided image-to-image translation problems is how the conditional guidance image is used to affect the processing of the input source image. 
Various forms of conditioning schemes have been proposed in the literature. 
The most common one is to directly concatenate the input source image and the guidance image at the input level (i.e., concatenation along the channel dimension). While being parameter efficient, this approach assumes that the additional guidance is required at the input level and the information can be carried through all the subsequent layers. Another commonly used  alternative is to concatenate the guidance and the input information at the feature level, assuming that the guidance feature representation is required at a certain level within the model.

A recent generalized conditioning scheme formalized as Feature-wise Linear Modulation (FiLM) has been successfully applied in visual reasoning task \cite{perez2018film}. In this scheme, affine transformations are applied to intermediate feature activations using scaling and shifting parameters learned from some external conditional information. In this approach, the learned scaling and shifting operations are applied \emph{feature-wise} (i.e., spatially invariant). 
There are other conditioning approaches similar to FiLM that have shown effectiveness in the context of style transfer. In this task, given an input image and a guidance style image, the goal is to synthesize an image that combines the content of the input image with the style of the guidance image.
One such approach is conditional instance normalization (CIN) \cite{dumoulin2017learned}, which can be seen as a FiLM layer replacing a normalization layer. In CIN, the feature representation is first normalized to zero mean and unit standard deviation. Then an affine transformation is applied to the normalized feature representation using scaling and shifting parameters learned from the guidance style image.
Another approach is adaptive instance normalization (AdaIN) \cite{huang2017arbitrary}. AdaIN is very similar to CIN, however, unlike CIN, it does not learn the affine transformation parameters but uses the mean and standard deviation of the guidance style image as the scaling and shifting parameters respectively. 

In this work, we propose a generalized conditioning scheme to incorporate the guidance image into the image-to-image translation model and show its applicability to different applications. There are two key differences between our proposed approach and the existing conditioning schemes. First, we propose to apply the conditioning operation in \emph{both} direction with information flowing not only from the guidance image to the input image, but from the input image to the guidance image as well. Second, we extend the existing feature-wise feature transformation to be \emph{spatially varying} to adapt to different contents in the input image. We refer to our proposed approach as bi-directional feature transformation (bFT). We validate the design of bFT through extensive experiments across multiple applications, including pose guidance appearance transfer, image synthesis with texture patch guidance, and joint depth upsampling. We demonstrate that our method, while not  application-specific, achieves competitive or better performance than the state-of-the-art. Through extensive ablation study, we also show that the proposed bFT is more effective than commonly used conditional schemes such as input/feature concatenation, CIN~\cite{dumoulin2017learned} and AdaIN~\cite{huang2017arbitrary}.

We make the following two contributions. First, we present the \emph{bi-directional} feature transformation for generic guided image-to-image translation tasks. Compared to existing approaches that only allow the information flow from guidance to the source image, we show that incorporating the information from the input to the guidance further help improve the performance of the end task. Second, we propose a \emph{spatially varying} extension of feature-wise transformation to better capture local contents from the guidance and the source image.

\begin{figure*}[t]    \centering
\mpage{0.24}{\includegraphics[width=0.7\linewidth]{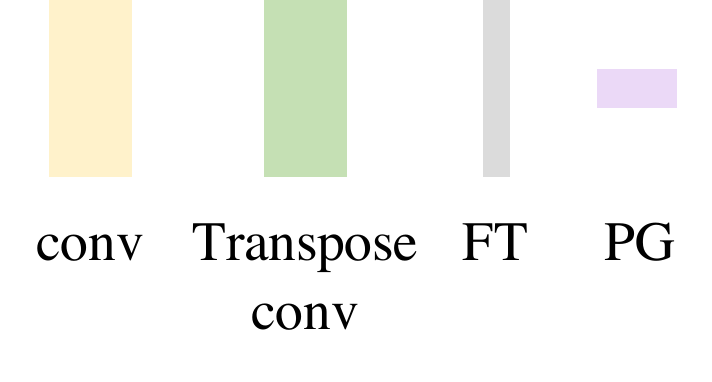} \includegraphics[width=\linewidth]{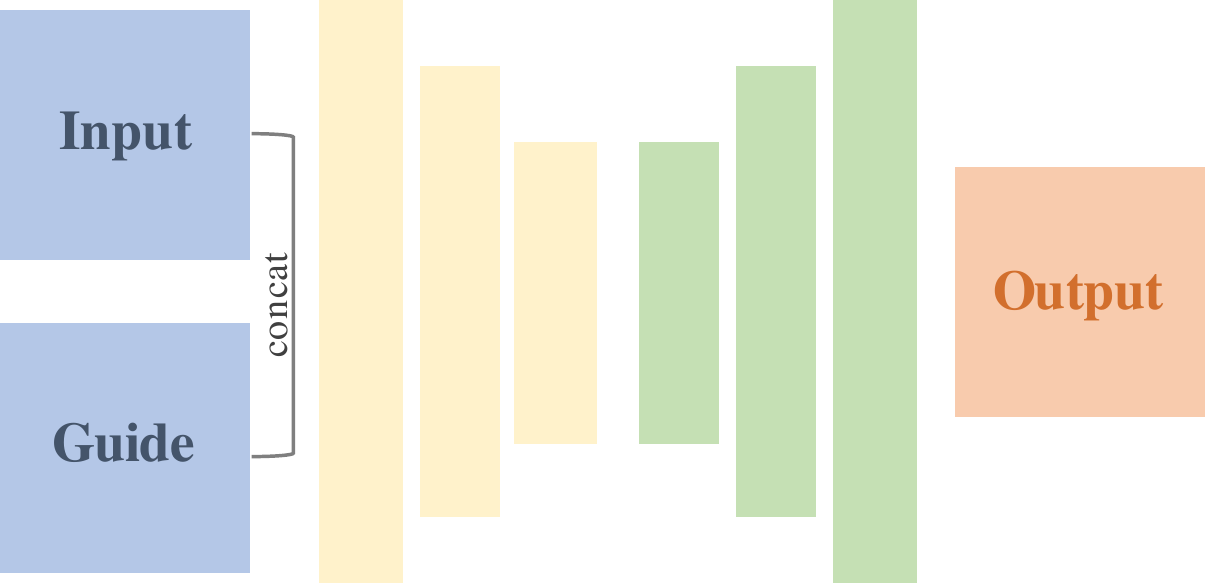}} \hfill
\mpage{0.24}{\includegraphics[width=\linewidth]{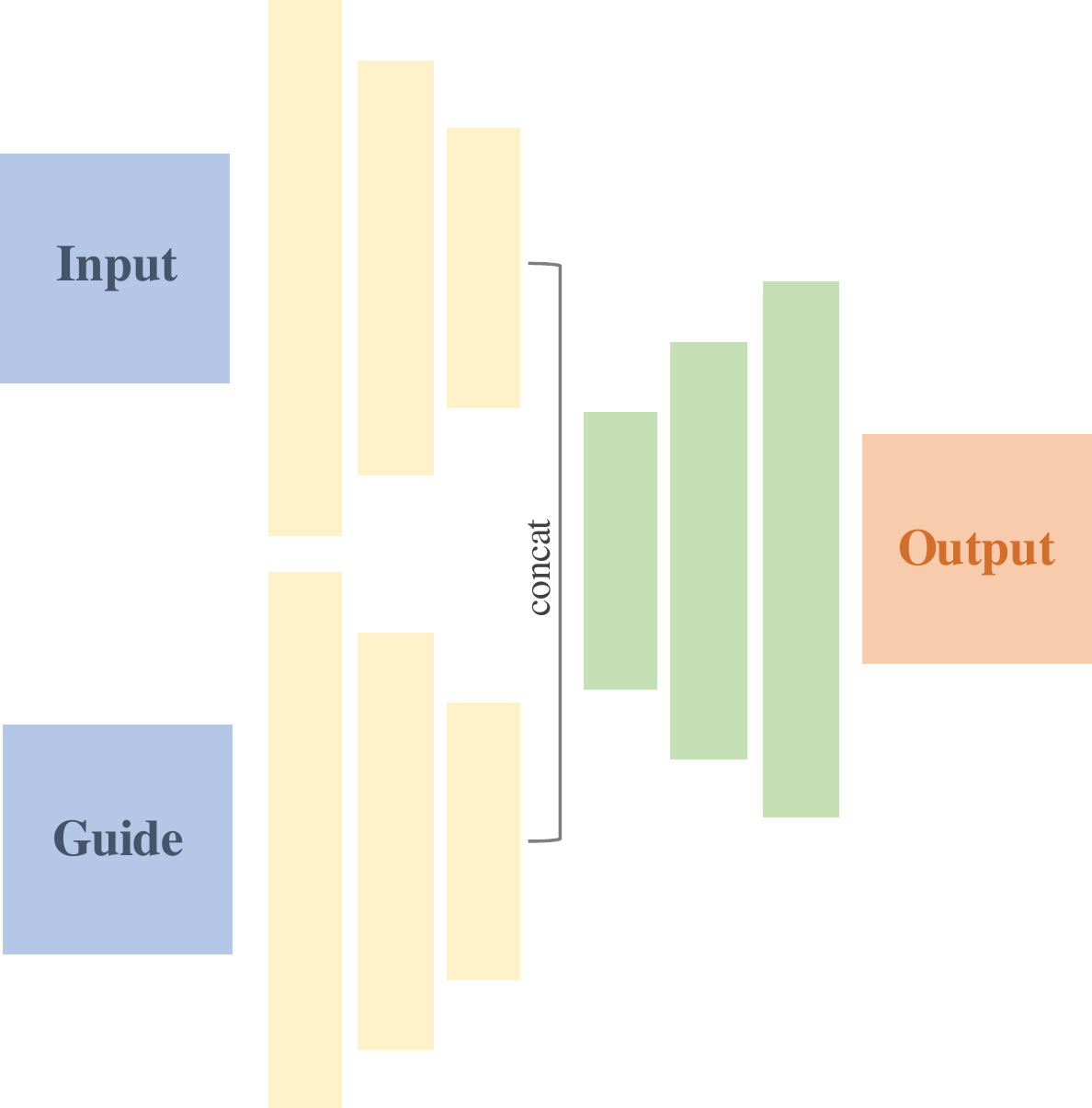}}\hfill
\mpage{0.24}{\includegraphics[width=\linewidth]{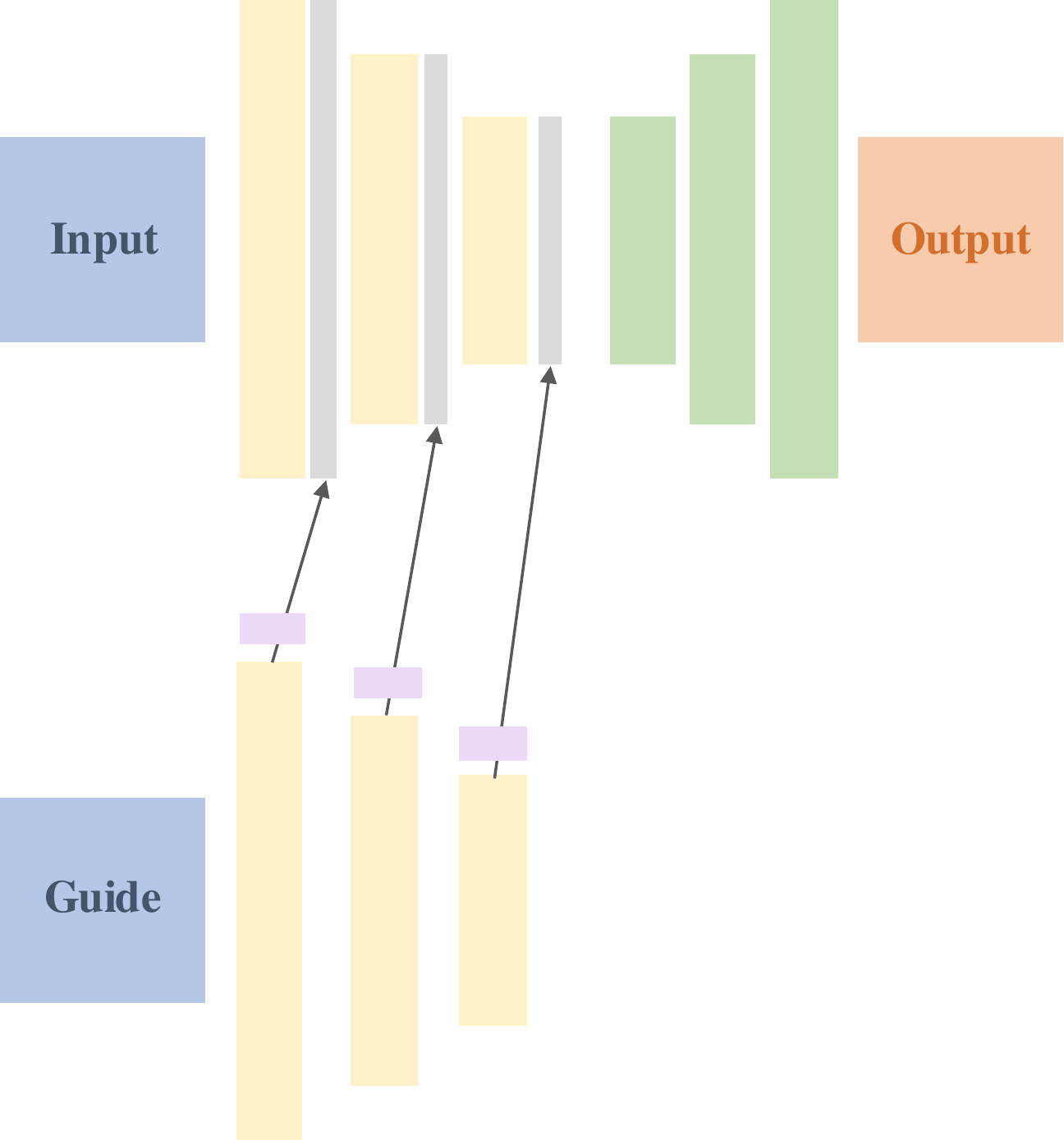}} \hfill
\mpage{0.24}{\includegraphics[width=\linewidth]{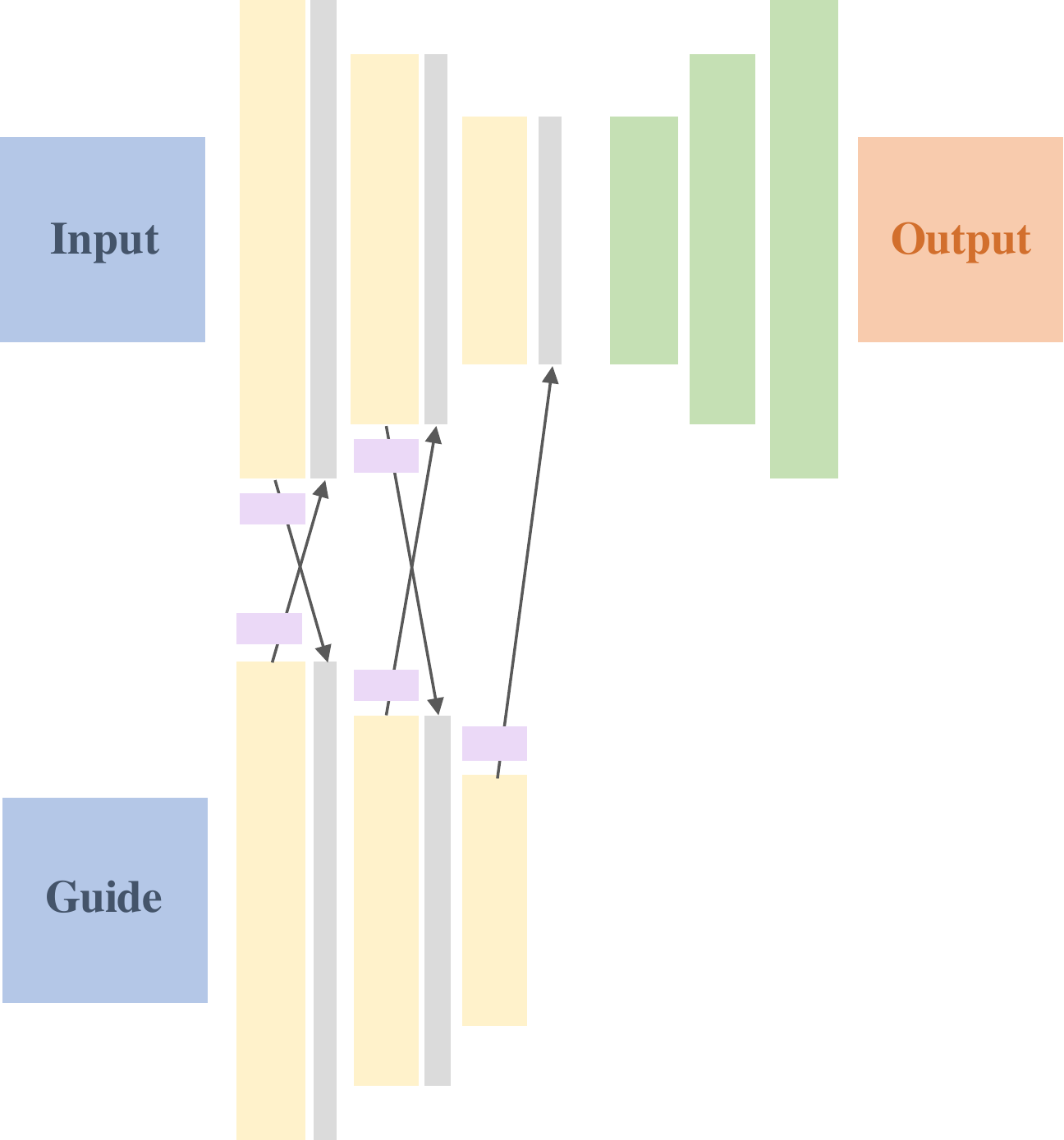}} \\
\vspace{1.0mm}
\mpage{0.24}{(a) Input Concatenation} \hfill
\mpage{0.24}{(b) Feature Concatenation}\hfill
\mpage{0.24}{(c) Uni-directional FT} \hfill
\mpage{0.24}{(d) Ours: Bi-directional FT} \\
\vspace{1.0mm}
\vspace{\figcapmargin}
\caption{\textbf{Conditioning schemes.} 
There are many schemes to incorporate the additional guidance into the image-to-image translation model. 
One straight forward scheme is (a) input concatenation, this will assume that we need the guidance image at the first stage of the model. 
Another scheme is (b) feature concatenation. It assumes that we need the feature representation of the guide before upsampling. 
In (c) we replace every normalization layer with our novel feature transformation (FT) layer that manipulates the input using scaling and shifting parameters generated from the guide using a parameter generator (PG). We denote this uni-directional scheme as uFT.
In this work, we propose (d) a bi-directional feature transformation scheme denoted as bFT. In bFT, the input is manipulated using scaling and shifting parameters generated from the guide and the guide is also manipulated using scaling and shifting parameters generated from the input.
}

\label{fig:motivation}
\end{figure*}

\vspace{\secmargin}
\section{Related Work}
\label{sec:related}

\vspace{\paramargin}
\paragraph{Image-to-image translation}
A generative model is an approach to learn a data distribution to generate new samples. One widely used technique is generative adversarial networks (GANs) \cite{goodfellow2014generative}. In GANs, there is a generator that tries to generate samples that look realistic to fool the discriminator, which tries to accurately tell whether a sample is real or fake. Conditional GANs extend the GANs by incorporating conditional information. One specific application of conditional GANs is image-to-image translation~\cite{isola2017image,wang2017high,park2019semantic}. Several recent advances include learning from unpaired dataset~\cite{zhu2017unpaired,yi2017dualgan,liu2017unsupervised}, improving diversity~\cite{lee2018diverse,huang2018multimodal,zhu2017toward}, application to domain adaptation~\cite{bousmalis2017unsupervised,hoffman2018cycada,chen2019crdoco}, and extension to video~\cite{wang2018video}.

Our work builds upon the recent advances in image-to-image translation and aims to extend it to a broader set of controllable image synthesis problems. We develop our network architecture similar to that of the pix2pix~\cite{isola2017image}, but the proposed bi-directional and spatially varying feature transformation layer is network-agnostic. 

\vspace{\paramargin}
\paragraph{Guided image-to-image translation}
A variant of image-to-image translation problem is to incorporate additional guidance image. In a guided image-to-image translation problem, we aim to translate an image from one domain into another while respecting certain constraints specified by a guidance image. This guidance image can take many forms. Examples include color strokes \cite{levin2004colorization, luan2007natural}, patches \cite{zhang2017real}, or color palette \cite{chang2015palette} to aid in user-guided colorization. The guidance can also be a domain label, as in a multi-domain image-to-image translation \cite{choi2017stargan}. Another form could be a style image as in the problem of style transfer \cite{dumoulin2017learned, ghiasi2017exploring, huang2017arbitrary}, a texture patch to texturize a sketch image \cite{xian2018texturegan}, or a high-resolution RGB image to aid in depth upsampling \cite{liu2018image,li2017joint}. Moreover, the guidance signal could be the multi-channel and sparse, such as pose landmark for pose guided person image synthesis problems \cite{ma2017pose,ma2018disentangled,siarohin2018deformable,neverova2018dense}. 
The guidance could also be a mask and sketch enabling users to inpaint and manipulate images \cite{yu2018free}. 
Due to the many different possible forms of the guidance images, most of the existing solutions for this class of problems are tailored toward specific applications, e.g., with specifically designed network architectures and training objectives.

Compared to many existing efforts in guided image-to-image translation, we focus on developing a conditioning scheme that is \emph{application-independent}. This makes our technique more widely applicable to many tasks with different forms of guidance.

\vspace{\paramargin}
\paragraph{Conditioning schemes}
\figref{motivation} compares with several commonly used conditioning schemes. The most straightforward way of performing guided image-to-image translation is to concatenate the input and the guidance image (along the feature channel dimension), followed by conventional image-to-image translation models. 
Such an input concatenation approach can be viewed as a simple conditioning scheme.
This approach assumes that the guidance signals are required from the input stage \cite{yu2018free,zhang2017real,xian2018texturegan}.
Several other types of conditioning schemes have been proposed in the literature.
Instead of concatenating the guidance and the input image at the input, one can also concatenate their feature activations at a certain layer~\cite{li2017joint,lai2018learning}. 
However, it may be non-trivial to choose a suitable level of the layer to concentrate input/guidance features for subsequent processing.
A recent and a more general scheme, Feature-wise Linear Modulation (FiLM) \cite{perez2018film}, applies feature-wise affine transformation using scaling and shifting parameters generated from conditioning information. 
Such a scheme has shown improved performance when applied to the problem of visual reasoning. Other variations of FiLM have shown good performance in the context of style transfer. Those approaches can be seen as replacing a normalization layer with a FiLM layer. One notable approach is the conditional instance normalization (CIN), where the scaling and shifting parameters are learned~\cite{dumoulin2017learned}. Another approach is adaptive instance normalization (AdaIN) where instead of learning the scaling and shifting parameters, the mean and standard deviation from the guidance features are used directly~\cite{huang2017arbitrary}.

Unlike existing conditioning schemes that allow information flow only from the guidance to the input (i.e., uni-directional conditioning), we show that the proposed \emph{bi-directional conditioning} method leads to sizable performance improvement. Furthermore, we generalize the existing spatially invariant feature-wise transform methods to support \emph{spatially varying} transformation.

\vspace{\secmargin}
\section{Bi-Directional Feature Transformation}
\label{sec:method}

\begin{figure*}[t]    \centering
\mpage{0.24}{\includegraphics[width=0.7\linewidth]{fig/legend.pdf}} \hfill
\mpage{0.73}{\includegraphics[width=\linewidth]{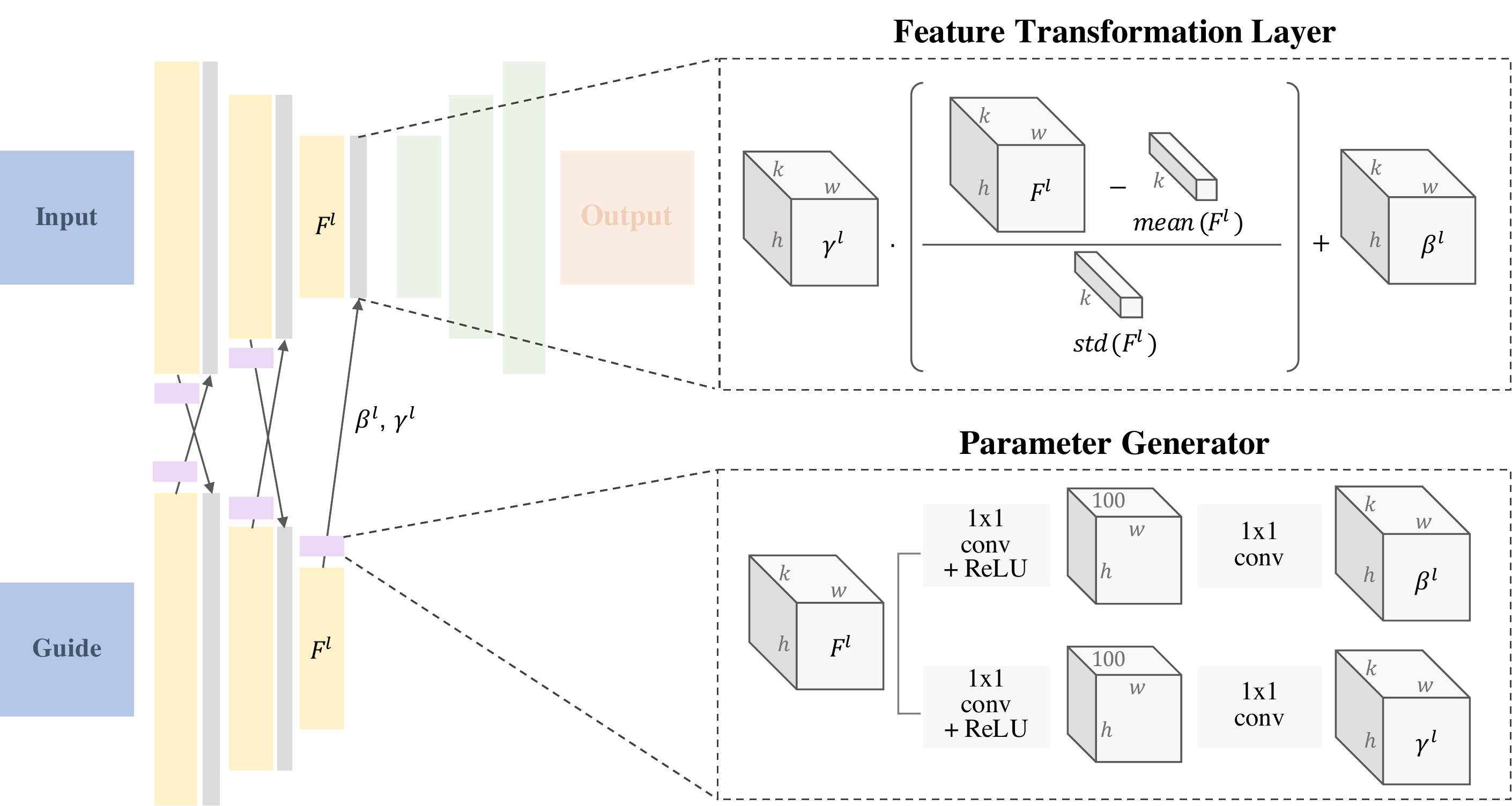}}
\\
\vspace{0.7mm}

\vspace{\figcapmargin}
\caption{\textbf{Bi-directional Feature Transformation.} 
We present a bi-directional feature transformation model to better utilize the additional guidance for guided image-to-image translation problems. 
In place of every normalization layer in the encoder, we add our novel FT layer. This layer scales and shifts the normalized feature of that layer as shown in \figref{tech_contrib}. 
The scaling and shifting parameters are generated using a parameter generation model of two convolution layers with a bottleneck of 100 dimension.
}

\label{fig:overview}
\end{figure*}

In this work, we aim to translate an image from one domain to another while respecting the constraints specified by a given guidance image. To tackle this problem, we propose Bi-Directional Feature Transformation (bFT) to incorporate the additional guidance image into the conditional generative model. We show that this conditioning scheme can be applied to various guided image-to-image translation problems without application-specific designs.

\begin{figure}[t]
\mfigure{0.35}{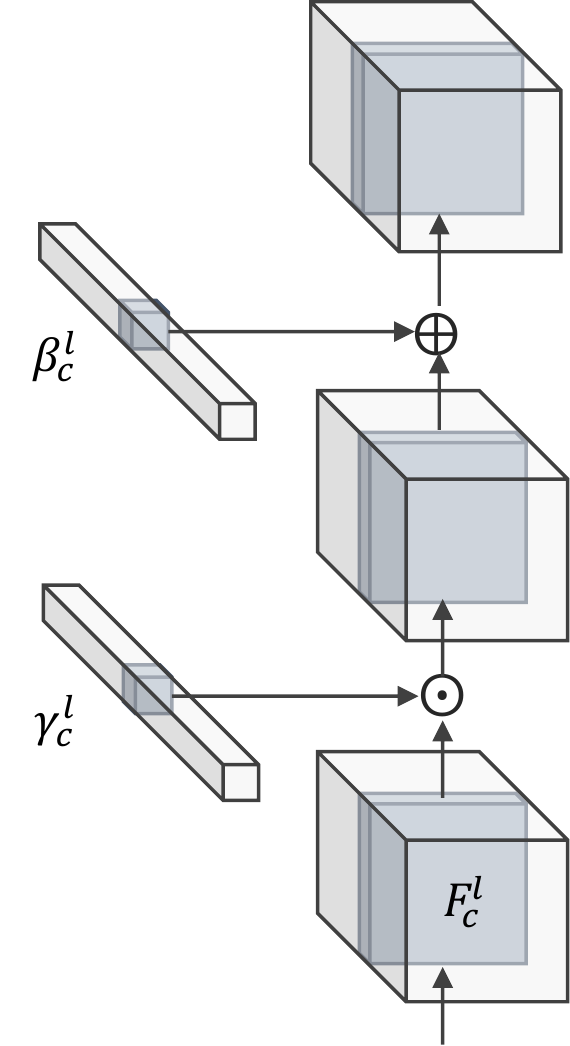} \hfill
\mfigure{0.45}{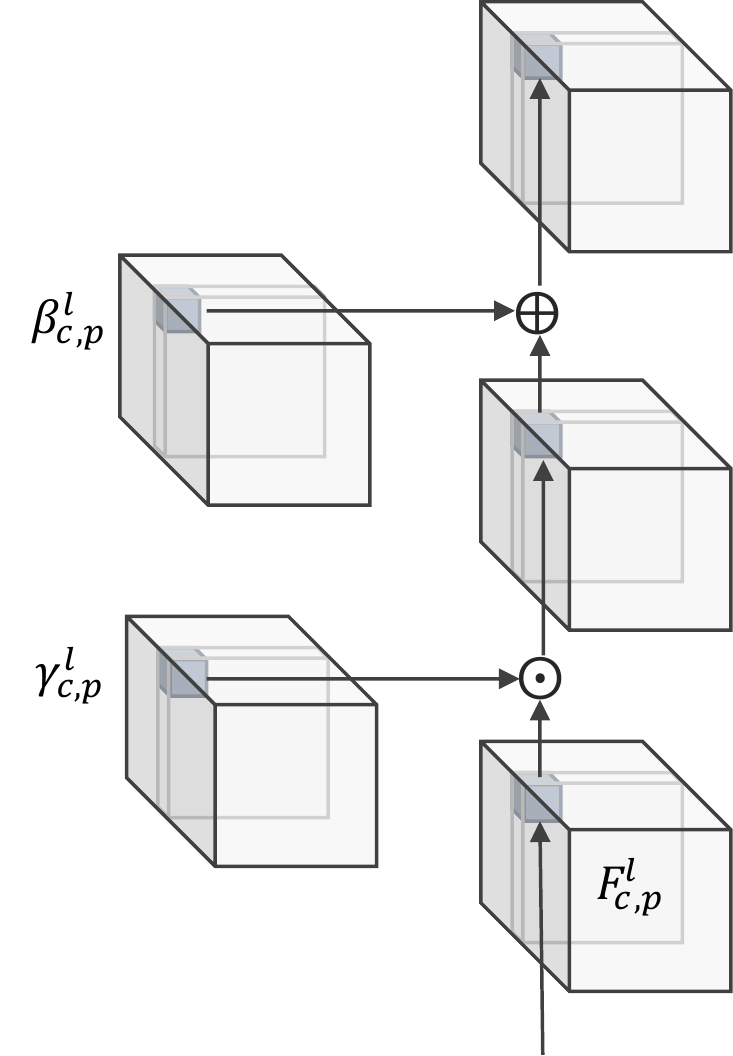} \\
\mpage{0.48}{(a) FiLM} \hfill
\mpage{0.48}{(b) FT (Ours)}\\
\caption{\textbf{Feature Transformation (FT).}
We present a feature transformation layer to incorporate the guidance into the image-to-image translation model.
A key difference between a FiLM layer and our FT layer is that the scaling $\gamma$ and shifting $\beta$ parameters of the FiLM layer are \emph{vectors}, while in our FT layer they are \emph{tensors}. Therefore, the scaling and shifting operations are applied in spatially varying manner in our FT layer in contrast to spatially invariant modulation as in the FiLM layer.
}

\label{fig:tech_contrib}
\end{figure}

\vspace{\secmargin}
\subsection{Feature transformation layer}
Here, we first present the feature transformation (FT) layer to incorporate the guidance information. In an FT layer, we perform an affine transformation on the normalized input features using scaling and shifting parameters computed from the features of the given guidance image. In Eqn.~\ref{eqn:FT}, we show this operation for an $l$-th layer. The scaling and shifting parameters $\gamma$ and $\beta$ are computed from the guidance signal using a \emph{parameter generator} shown in \figref{overview}.

\begin{equation}\label{eqn:FT}
F_\mathrm{input}^{l+1} = \gamma_\mathrm{guide}^{l} \frac{F_\mathrm{input}^{l} - \mathrm{mean}(F_\mathrm{input}^{l})}{\mathrm{std}(F_\mathrm{input}^{l})} + \beta_\mathrm{guide}^{l}.
\end{equation}

A key difference between the FiLM layer \cite{perez2018film} and the proposed FT layer is highlighted in \figref{tech_contrib}. Specifically, the scaling $\gamma$ and shifting $\beta$ parameters of the FiLM layers are \emph{vectors} and are applied channel-wise. That is, the same affine transformation of feature activations is applied the same way regardless of the spatial position on the feature map. Such approaches are reasonable for tasks such as style transfer or visual reasoning. However, they may not be able to capture fine-grained spatial details that are important for image-to-image translation problems. In contrast, the parameters in our FT layer are three-dimensional \emph{tensors} which offer a flexible way for modulating the input features in a spatially varying manner and supports various forms of guidance signals (e.g., dense, sparse, or multi-channel).

\vspace{\secmargin}
\subsection{Bi-directional conditioning scheme}
To further utilize the available information from the guidance image, we propose a \emph{bi-directional conditioning scheme}. 
Unlike existing conditioning schemes that only allow the guidance signal to influence the input image process, our approach supports bi-directional communication between two branches of the networks processing the input and guidance image.
This bi-directional flow of information enables the generative model to better capture the constraints of the guidance image.
In our proposed bFT scheme, we replace every normalization layer with our proposed FT layer. At $l$-th layer, the guidance feature representation manipulates the input feature representation as shown in Eqn.~\ref{eqn:FT}, and at the same time is manipulated by that input feature representation. Such that:
\begin{equation}\label{eqn:FT_}
F_\mathrm{guide}^{l+1} = \gamma_\mathrm{input}^{l} \frac{F_\mathrm{guide}^{l} - \mathrm{mean}(F_\mathrm{guide}^{l})}{\mathrm{std}(F_\mathrm{guide}^{l})} + \beta_{input}^{l}
\end{equation}
Our intuition is that such a bi-directional approach can be seen as a bi-directional communication between a teacher (guidance branch) and a student (input image branch). 
A one-way communication from the teacher to the student might not help the student understand the teacher as much as two-way communication. 

\vspace{\secmargin}
\section{Experimental Results}
\label{sec:results}
\begin{figure*}[hbtp]
\centering

\mpage{0.01}{\raisebox{0pt}{\rotatebox{90}{Input/Guide}}} 
\mfigure{0.09}{{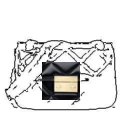}}\hfill
\mfigure{0.09}{{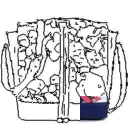}}\hfill
\mfigure{0.09}{{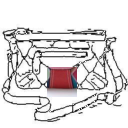}}\hfill
\mfigure{0.07}{{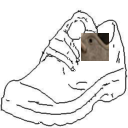}}\hfill
\mfigure{0.07}{{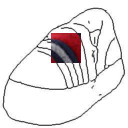}}\hfill
\mfigure{0.07}{{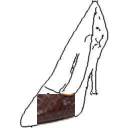}}\hfill
\mfigure{0.07}{{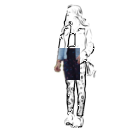}}\hfill
\mfigure{0.07}{{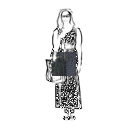}}\hfill
\mfigure{0.07}{{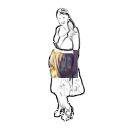}}\hfill

\mpage{0.01}{\raisebox{0pt}{\rotatebox{90}{pix2pix}}} 
\mfigure{0.09}{{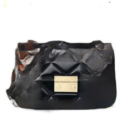}}\hfill
\mfigure{0.09}{{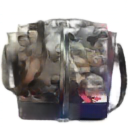}}\hfill
\mfigure{0.09}{{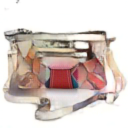}}\hfill
\mfigure{0.07}{{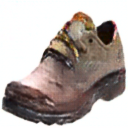}}\hfill
\mfigure{0.07}{{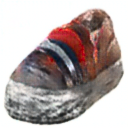}}\hfill
\mfigure{0.07}{{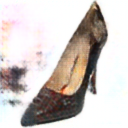}}\hfill
\mfigure{0.07}{{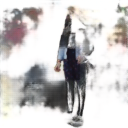}}\hfill
\mfigure{0.07}{{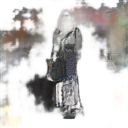}}\hfill
\mfigure{0.07}{{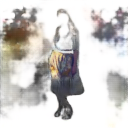}}\hfill

\mpage{0.01}{\raisebox{0pt}{\rotatebox{90}{Xian \etal}}} 
\mfigure{0.09}{{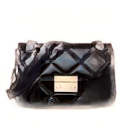}}\hfill
\mfigure{0.09}{{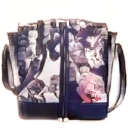}}\hfill
\mfigure{0.09}{{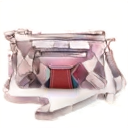}}\hfill
\mfigure{0.07}{{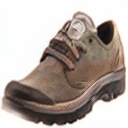}}\hfill
\mfigure{0.07}{{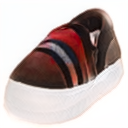}}\hfill
\mfigure{0.07}{{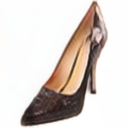}}\hfill
\mfigure{0.07}{{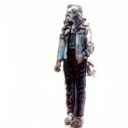}}\hfill
\mfigure{0.07}{{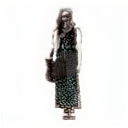}}\hfill
\mfigure{0.07}{{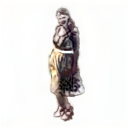}}\hfill

\mpage{0.01}{\raisebox{0pt}{\rotatebox{90}{Ours}}} 
\mfigure{0.09}{{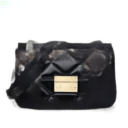}}\hfill
\mfigure{0.09}{{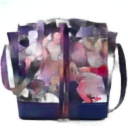}}\hfill
\mfigure{0.09}{{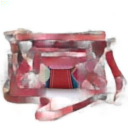}}\hfill
\mfigure{0.07}{{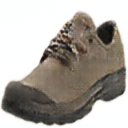}}\hfill
\mfigure{0.07}{{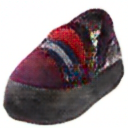}}\hfill
\mfigure{0.07}{{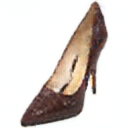}}\hfill
\mfigure{0.07}{{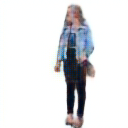}}\hfill
\mfigure{0.07}{{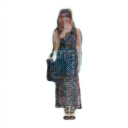}}\hfill
\mfigure{0.07}{{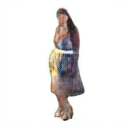}}\hfill

\mpage{0.01}{\raisebox{0pt}{\rotatebox{90}{Target}}} 
\mfigure{0.09}{{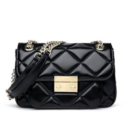}}\hfill
\mfigure{0.09}{{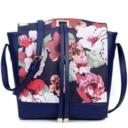}}\hfill
\mfigure{0.09}{{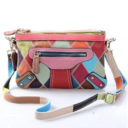}}\hfill
\mfigure{0.07}{{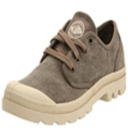}}\hfill
\mfigure{0.07}{{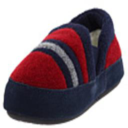}}\hfill
\mfigure{0.07}{{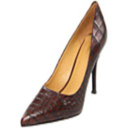}}\hfill
\mfigure{0.07}{{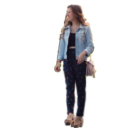}}\hfill
\mfigure{0.07}{{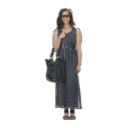}}\hfill
\mfigure{0.07}{{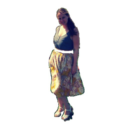}}\hfill

\vspace{\figcapmargin}
\captionof{figure}{\textbf{Controllable sketch-to-photo synthesis with texture patches.} Texture transfer qualitative comparison with state-of-the-art-results on the handbags, shoes, and clothes datasets from \cite{xian2018texturegan}. Here we use the ground truth texture patches as the guidance signal. 
}
\label{fig:texture}
\end{figure*}

We evaluate our proposed bi-directional feature transformation conditioning scheme on three different guided image-to-image translation problems with three different types of the guidance signal.\footnote{Code available: \url{https://github.com/vt-vl-lab/Guided-pix2pix}} 
For all tasks, we use GANs with two possible architectures as our generator model, either Unet or Resnet. 
We follow the same training objective function (a weighted combination of $L_1$ loss and an adversarial loss $L_{GAN}$) as in \cite{isola2017image}:
\begin{equation}\label{eqn:obj}
L_{GAN} (G, D) + \lambda L_1(G).
\end{equation}
where we set $\lambda$ to 100 for all the experiments.
For each task we compare our results with state-of-the-art methods as well as pix2pix \cite{isola2017image} (with input concatenation conditioning).

\vspace{\secmargin}
\subsection{Controllable sketch-to-photo synthesis}
In this texture transfer task, given a sketch and a random sized texture patch as the guidance signal, we aim to synthesize a photo that fills the input sketch respecting that given texture patch. 

\vspace{\paramargin}\paragraph{Implementation details} We use the Unet architecture of \cite{isola2017image} as the base architecture of our model. For both our bFT model and pix2pix, we train using a learning rate of 0.0002 with 7 layers of Unet architecture. We use an Adam optimizer for both with beta1 as 0.5 for pix2pix, and beta1 as 0.9 for our model.
For the handbag dataset, we train for 500 epochs with a batch size of 64. For the shoes and clothes datasets, we train for 100 epochs with batch size of 256.

\vspace{\paramargin}\paragraph{Datasets and metrics} We use the 128x128 data generated by Xian \etal\cite{xian2018texturegan} and follow the same texture patch generation algorithm from the ground truth images. We evaluate the results using the Learned Perceptual Image Patch Similarity (LPIPS) metric proposed by Zhang \etal\cite{zhang2018perceptual} and the frechet inception distance (FID) proposed by Heusel \etal\cite{heusel2017gans}. For every sketch in the test set, we generate 10 random sized ground truth texture patches using the texture patch generation algorithm from Xian \etal\cite{xian2018texturegan} and compute the LPIPS and the FID of the synthesized images. We use the provided pretrained models of Xian \etal\cite{xian2018texturegan} to compute their results. Their pretrained models are trained on ground truth patches as well as external patches, while our model and pix2pix are trained only on ground truth patches.

\vspace{\paramargin} \paragraph{Evaluation}
We show the quantitative results of our work compared to Isola \etal\cite{isola2017image} and Xian \etal\cite{xian2018texturegan} in Table \ref{tab:texture}. While our model training is considerably simpler (trained with only two losses) than that of the Xian \etal\cite{xian2018texturegan} (with seven different loss terms), we show favorable results against both  pix2pix~\cite{isola2017image} and Xian \etal\cite{xian2018texturegan} in terms of the LPIPS metric on all three datasets. We also show the FID results.

We show sample qualitative results on the handbag, shoes, and clothes datasets in \figref{texture} using ground truth texture patches as the guidance signal. %

\begin{table}[t]\setlength{\tabcolsep}{2pt}
	\centering\footnotesize
	\caption{Texture Transfer Task: visual quality evaluation using the Learned Perceptual Image Patch Similarity (LPIPS) metric \cite{zhang2018perceptual} and Frechet Inception Distance (FID) \cite{heusel2017gans} on the datasets generated by \cite{xian2018texturegan}. A lower score is better.}
	\begin{tabular}{ccccccc}
		\toprule
		        &  \multicolumn{2}{c}{Handbag Dataset} & \multicolumn{2}{c}{Shoes Dataset} & \multicolumn{2}{c}{Clothes Dataset}\\
		        & LPIPS & FID & LPIPS & FID & LPIPS & FID\\
		\midrule
        Xian \etal\cite{xian2018texturegan} & 0.171 & 60.848 & 0.124 &44.762 &0.113 & 49.568\\
        \midrule
        pix2pix \cite{isola2017image}       & 0.234 & 96.31 & 0.238 & 197.492&0.439 & 190.161\\
        Ours        & 0.161 & 74.885 &0.124 & 121.241 & 0.067  & 58.407\\
        \bottomrule
		\vspace{\tabmargin}
		\label{tab:texture}
	\end{tabular}
\end{table}

\begin{figure*}[t]
\centering
\mpage{0.11}{Input} \hfill
\mpage{0.11}{Guide}\hfill
\mpage{0.11}{Ma \cite{ma2017pose}} \hfill
\mpage{0.11}{Siarohin \cite{siarohin2018deformable}} \hfill
\mpage{0.11}{pix2pix \cite{isola2017image}} \hfill
\mpage{0.11}{Ours} \hfill
\mpage{0.11}{Target} \\
\mfigure{0.11}{{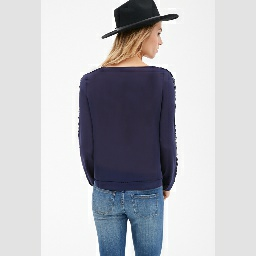}}\hfill
\mfigure{0.11}{{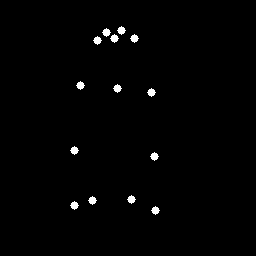}}\hfill
\mfigure{0.11}{{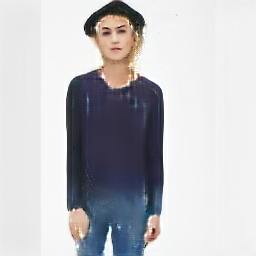}}\hfill
\mfigure{0.11}{{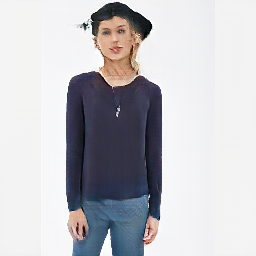}}\hfill
\mfigure{0.11}{{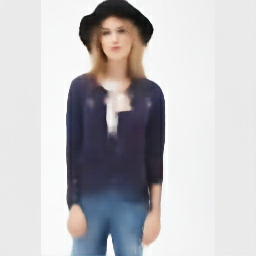}}\hfill
\mfigure{0.11}{{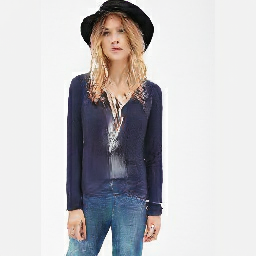}}\hfill
\mfigure{0.11}{{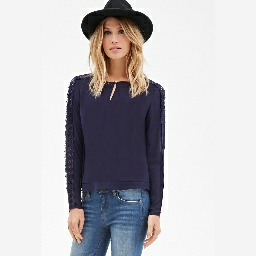}} \\
\mfigure{0.11}{{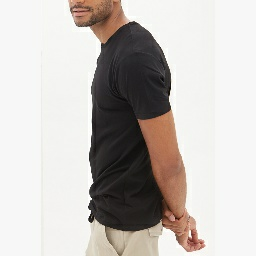}}\hfill
\mfigure{0.11}{{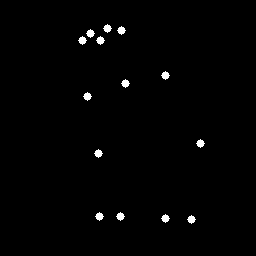}}\hfill
\mfigure{0.11}{{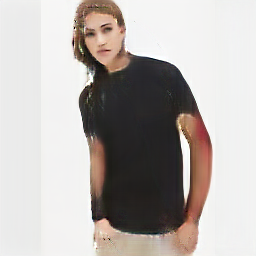}}\hfill
\mfigure{0.11}{{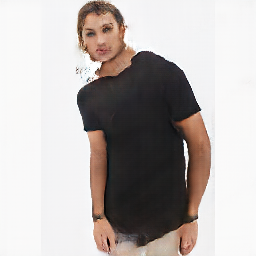}}\hfill
\mfigure{0.11}{{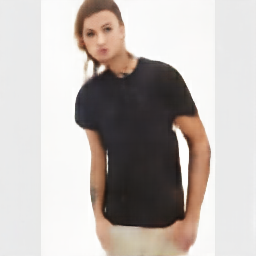}}\hfill
\mfigure{0.11}{{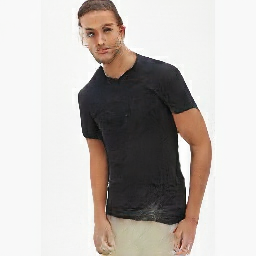}}\hfill
\mfigure{0.11}{{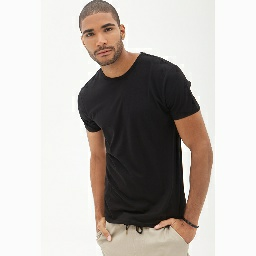}} \\
\mfigure{0.11}{{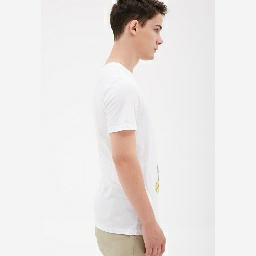}}\hfill
\mfigure{0.11}{{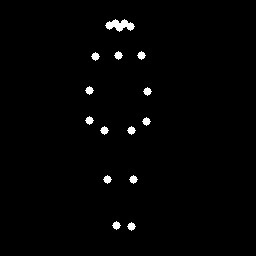}}\hfill
\mfigure{0.11}{{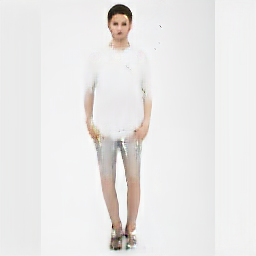}}\hfill
\mfigure{0.11}{{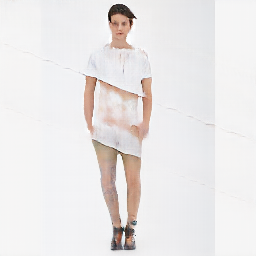}}\hfill
\mfigure{0.11}{{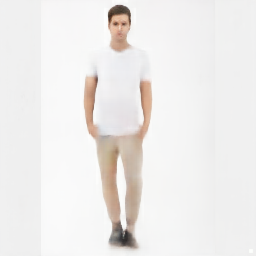}}\hfill
\mfigure{0.11}{{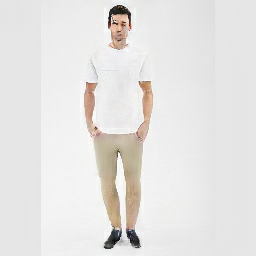}}\hfill
\mfigure{0.11}{{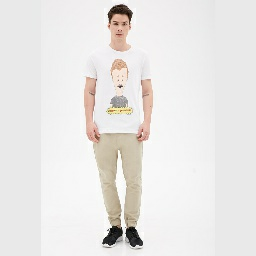}} \\

\vspace{\figcapmargin}
\captionof{figure}{\textbf{Controllable person-image synthesis with pose keypoints.} Pose transfer qualitative results on DeepFashion dataset. Our model in general achieves sharper results on this challenging task. 
}
\label{fig:pose}
\end{figure*}

\vspace{\secmargin}
\subsection{Controllable person-image synthesis}
In the pose transfer task, given an image of a person and a target pose as a guidance signal, we aim to synthesize an image of that given person in the desired pose.

\vspace{\paramargin} \paragraph{Implementation details} We use ResNet architecture as the base architecture of our model. For both our bFT model and pix2pix, we train for 100 epochs using a learning rate of 0.0002 with a batch size of 8, then we minimize the learning rate to 0.00002 and train for 50 additional epochs. We use the Adam optimizer for both with beta1 as 0.5 for pix2pix, and beta1 as 0.9 for our model. 
We use 8 layers for the Unet architecture for pix2pix.

\vspace{\paramargin} \paragraph{Datasets and metrics} We use the 256x256 train and test sets provided by Ma \etal\cite{ma2017pose} from the DeepFashion dataset \cite{liu2016deepfashion}. Following the evaluation protocols in literature, we use both SSIM and Inception Score (IS) to measure the quality of the synthesized images. We also use the FID metric.

\vspace{\paramargin} \paragraph{Evaluation}
We show the quantitative results of our work compared to state-of-the-art methods in Table \ref{tab:pose}. We note that Siarohin \etal\cite{siarohin2018deformable} trains on a different training set of the DeepFashion dataset and excludes samples where pose keypoints are not detected. To ensure fair comparison, we modify our test set to exclude such samples. We report the results on both the full test set and the modified one. We use the pretrained models provided by \cite{siarohin2018deformable,ma2017pose} to test their models on our test set. We also note that Siarohin \etal\cite{siarohin2018deformable} uses the input pose as an additional input to the model. We show favorable results against other methods using the Frechet Inception Distance (FID).

Note that it is very difficult to measure the quality of a synthesized image. In this task, however, we not only care about the quality of the image, but also about it having the same content and respecting the target pose. 
We show the qualitative results in \figref{pose}. 

Unlike the aforementioned methods that use keypoint based pose, Neverova \etal\cite{neverova2018dense} uses dense pose to perform pose transfer and achieved a score of [SSIM=0.785, IS=3.61], however, we were unable to obtain the data nor the pre-trained model for comparison.

\begin{table}[t]\setlength{\tabcolsep}{4pt}
	\centering\footnotesize
	\caption{Pose Transfer task: visual quality evaluation on the DeepFashion dataset \cite{liu2016deepfashion}. A higher score of SSIM/IS is better. A lower score of FID is better.}
	\begin{tabular}{ccccccc}
		\toprule
    	 &   \multicolumn{3}{c}{Full test set} & \multicolumn{3}{c}{Modified test set} \\
		  &  SSIM & IS & FID & SSIM & IS &FID \\
		\midrule
        Ma \etal\cite{ma2018disentangled}               & 0.614 & \underline{3.29} & - & - & - & -\\
        Ma \etal\cite{ma2017pose}                       & 0.762 & 3.09   & 47.917 & 0.764 & 3.10 & 47.373\\
        Siarohin \etal\cite{siarohin2018deformable}     & 0.758 & \textbf{3.36} & \underline{15.655} & 0.763 & \textbf{3.32} & \underline{15.215}\\
        \midrule
        pix2pix \cite{isola2017image}           & \textbf{0.770} & 2.96 & 66.752 & \textbf{0.774} & 2.93 & 65.907\\
        Ours    & \underline{0.767} & 3.22    & \textbf{12.266} & \underline{0.771} & \underline{3.19} & \textbf{12.056}\\
        \bottomrule
		\vspace{\tabmargin}
		\label{tab:pose}
	\end{tabular}
\end{table}

\begin{figure*}[t]
\centering

\mpage{0.13}{Input} \hfill
\mpage{0.13}{Guide}\hfill
\mpage{0.13}{DJF }\hfill
\mpage{0.13}{DJFR }\hfill
\mpage{0.13}{pix2pix } \hfill
\mpage{0.13}{Ours} \hfill
\mpage{0.13}{Target} \\

\mfigure{0.99}{{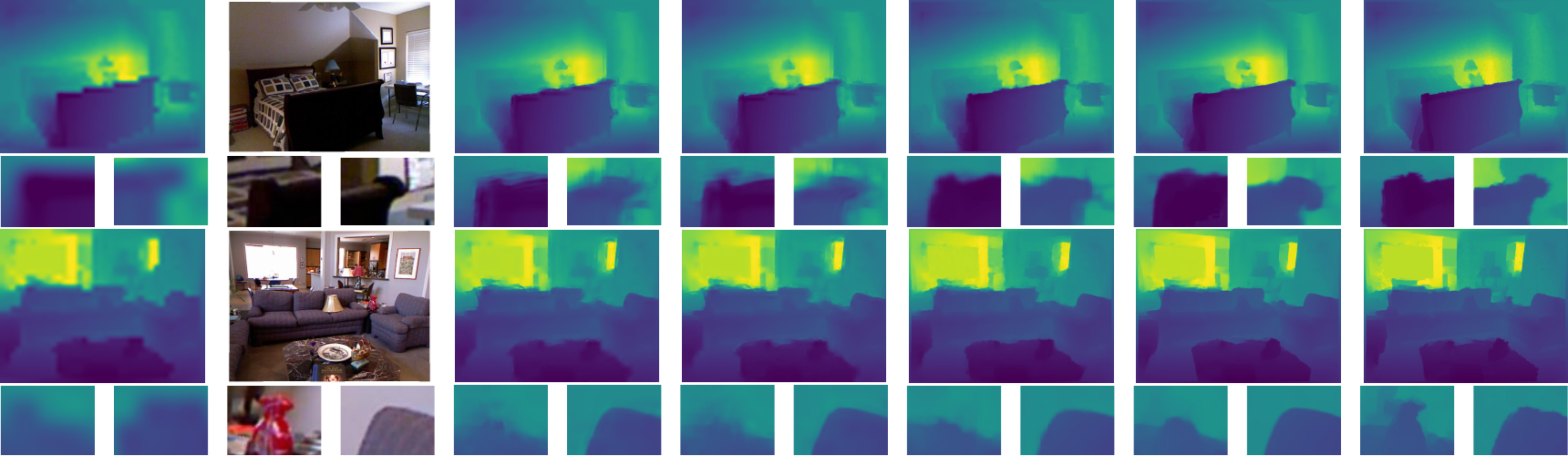}}\hfill

\vspace{\figcapmargin}
\captionof{figure}{\textbf{Depth upsampling guided by an RGB image.} Comparison of depth upsampling qualitative results for a scale factor of 16 with the state-of-the-art methods. The zoomed-in crops show that our method is able to capture fine details with sharper edges.
}
\label{fig:depth}
\end{figure*}

\vspace{\secmargin}
\subsection{Depth upsampling}
In depth upsampling, we aim to generate a high-resolution depth map given a low resolution depth map with the guidance of a high resolution RGB image. 

\vspace{\paramargin} \paragraph{Implementation details} We use the ResNet architecture as the base architecture of our model. For both our bFT model and pix2pix, we only use L1 as the objective function and train for 500 epochs using a learning rate of 0.0002 with batch size of 2. We use an Adam optimizer for both with beta1 as 0.5. For our work, we train on the original size of the data 480x640, however, because pix2pix uses square sized inputs, it is trained on 512x512 resized data and we resize back before evaluation. We use 9 layers for the Unet architecture of pix2pix.

\vspace{\paramargin} \paragraph{Dataset and metric} Following the setting of Li \etal\cite{li2017joint}, we use 1000 samples from the NYU v2 dataset \cite{NYUdataset} for training and we test on the remaining 449. We generate the low resolution input depth map using bicubic upsampling for three different scale factors 16, 8, and 4.
Similar to the works in literature we use RMSE to evaluate the quality of the generated depth.

\vspace{\paramargin} \paragraph{Evaluation}
We show the RMSE results of our work compared to Isola \etal\cite{isola2017image} and state-of-the-art methods in Table \ref{tab:depth}. We report the results by Li \etal\cite{li2017joint}. We also show qualitative results for the three scale factors in \figref{depth}. Our model, while not designed for depth upsampling, can achieve state-of-the-art performance.

\begin{table}[t]\setlength{\tabcolsep}{10pt}
	\centering\footnotesize
	\caption{Depth Upsampling task: root mean square error (RMSE) results in centimeters for the NYU v2 dataset \cite{NYUdataset}.}
	\begin{tabular}{cccc}
		\toprule
		Depth Scale &  $x4$ & $x8$ & $x16$ \\
		\midrule
		Bicubic& 8.16 & 14.22 & 22.32 \\
		MRF \cite{diebel2006application}& 7.84 & 13.98 & 22.20\\
		GF \cite{he2013guided}& 7.32 & 12.98 & 22.03 \\
		JBU \cite{kopf2007joint}& 4.07 & 13.62 & 22.03 \\
		Ham \cite{ham2015robust}& 5.27 & 12.31 & 19.24 \\
		DMSG \cite{hui2016depth}& 3.48 & 6.07 & 10.27 \\
		FBS \cite{barron2016fast}& 4.29 & 8.94 & 14.59 \\
		DJF \cite{li2016deep} & 3.54 & 6.20 & 10.21 \\
        DJFR \cite{li2017joint}& \underline{3.38} & \underline{5.86} & \underline{10.11} \\
        \midrule
        pix2pix \cite{isola2017image}   & 4.12 & 6.48 & 10.17 \\
        Ours &  \textbf{3.35} & \textbf{5.73} & \textbf{9.01}\\
        \bottomrule
		\vspace{\tabmargin}
		\label{tab:depth}
	\end{tabular}
\end{table}

\begin{table*}[htbp]\setlength{\tabcolsep}{3pt}
	\centering\footnotesize
    \caption{Conditioning schemes. }\label{tab:conditioning}
    \begin{tabular}{l|cccccccccccc}
		\toprule
		Conditioning method & \multicolumn{3}{c}{Depth Upsampling}  & \multicolumn{3}{c}{Pose Transfer} & \multicolumn{6}{c}{Texture Transfer}\\
		& & & & & & & \multicolumn{2}{c}{Handbags} & \multicolumn{2}{c}{Shoes} & \multicolumn{2}{c}{Clothes} \\
		                    & 4x &  8x & 16x                        & SSIM &  IS & FID & LPIPS & FID & LPIPS & FID & LPIPS & FID\\
		\midrule
        Input Concatenation     & 6.65 & 8.42 & 11.86               & 0.782 &  3.10 & 42.330 & 0.182 & 85.600 & 0.137 & 124.973 & 0.061 & 60.795\\
        Feature Concatenation   & 6.67 & 7.63 & 11.59               & 0.770 &  3.26 & 14.672 & 0.196 & 87.052 & 0.145 & 104.227 & 0.085 & 44.900\\
        uFT                     & 5.55 & 7.26 & 11.41               & 0.765 &  3.18 & 13.988 & 0.174 & 85.273 & 0.126 & 119.588 & 0.071 & 56.66\\
        bFT (Ours)              & 3.35 & 5.73 & 9.01                & 0.767 &  3.17 & 13.240 & 0.171 & 80.179 & 0.123 & 119.832 & 0.067 & 58.467\\
        \bottomrule
	\end{tabular}
\vspace{\tabmargin}
\end{table*}%

\begin{table}[htbp]\setlength{\tabcolsep}{2pt}
	\centering\footnotesize
    \caption{Number of feature transformation (FT) layers.}\label{tab:layers}
    \begin{tabular}{l|cccccccc}
		\toprule
		\#Layers & \multicolumn{2}{c}{Depth Upsampling} & \multicolumn{6}{c}{Pose Transfer}\\
		       & uFT & bFT &\multicolumn{3}{c}{uFT}&  \multicolumn{3}{c}{bFT}\\
		       &  x16   &  x16   &     SSIM & IS & FID  &     SSIM & IS & FID \\
		\midrule
        1  & 10.79 & 10.79              & 0.786 & 2.92 & 59.678 &  0.786 & 2.92 & 59.678\\
        2 & 10.75 & 8.96                & 0.784 & 2.98 & 47.411 &  0.785 & 3.01 & 51.458\\
        3 & 10.26 & 8.82                & 0.768 & 3.15 & 16.069&  0.766 & 3.24 & 13.392\\
        4 & 11.41 & 9.01                & 0.765 & 3.18 & 13.988 &  0.767 & 3.17 & 13.240\\
        \bottomrule
	\end{tabular}
	\vspace{\tabmargin}
  \end{table}

\begin{table}[htbp]\setlength{\tabcolsep}{4pt}
	\centering\footnotesize
    \caption{Different approaches to affine transformation.}\label{tab:affine}
    \begin{tabular}{l|cccc}
		\toprule
		Method & Depth Upsampling  & \multicolumn{3}{c}{Pose Transfer} \\
         	                  & x16               &  SSIM & IS & FID \\
		\midrule
        Ours                     & 9.01              & 0.767 & 3.17 &
         13.240 \\
        bi-directional AdaIN                 & 13.36             & 0.722 & 3.37 & 160.846 \\
        bi-directional CIN                   & 13.97             & 0.721 & 3.36 & 157.335 \\
        \midrule
        Final Layer - FT                      & 11.40              & 0.769 & 3.25 & 18.292 \\
        Final Layer - AdaIN                   & 14.30              & 0.720 & 3.30 & 146.596 \\
        Final Layer - CIN                     & 14.51              & 0.720 & 3.58 & 168.503 \\
        \bottomrule
	\end{tabular}
	\vspace{\tabmargin}
\end{table}

\vspace{\secmargin}
\subsection{Ablation study}
We conduct an ablation study to the effectiveness of our proposed bi-directional conditioning scheme.

\vspace{\paramargin} \paragraph{Conditioning schemes}
We compare our proposed bi-directional feature transformation scheme (bFT) to uni-directional feature transformation (uFT), feature concatenation, and input concatenation schemes shown in \figref{motivation}. We show quantitative results in Table \ref{tab:conditioning}.

\vspace{\paramargin} \paragraph{Number of feature transformation (FT) layers}
In our bFT model, we use FT in place of every normalization layer. For pose transfer and depth upsampling tasks, we use a Resnet base with 4 normalization layers. Replacing those layers with our proposed FT layer, we end up with 4 FT layers. We compare our approach with using FT at l, 2, and 3 layers both bi-directionally and uni-directionally. We show the quantitative results in Table \ref{tab:layers}.

\vspace{\paramargin} \paragraph{Different approaches to affine transformation}
Using our bi-directional approach, we compare our proposed FT with CIN and AdaIN. In both CIN and AdaIN, we use FiLM layer in place of every normalization layer. In CIN, we learn the scaling and shifting parameters, while in AdaIN, we use the mean as the scaling parameter and the standard deviation as the shifting parameter. 
We also test feature transformation at only the last layer of the encoder and compare the performance of our FT with CIN and AdaIN. 
We show the quantitative results in Table \ref{tab:affine}.

\ignorethis{
\vspace{\paramargin} \paragraph{bFT location}
We test different variations of where to apply our proposed bFT. In our work, we apply bFT between the encoder of the input and the encoder of guide. We compare our approach with bFT between the decoder of the input and the decoder of the guide, as well as between the decoder of the input and the encoder of the guide.  The quantitative results are shown in Table \ref{tab:location}.
\begin{table}[h]\setlength{\tabcolsep}{5pt}
	\centering\footnotesize
    \caption{bFT location.}\label{tab:location}
    \begin{tabular}{l|ccc}
		\toprule
		Method       & Depth Upsampling  & \multicolumn{2}{c}{Pose Transfer}\\
        guide-input                 & x16               &  SSIM & IS\\
		\midrule
		Ours              & 9.01              & 0.767 & 3.17\\
        decoder-decoder          & 10.97            & 0.760 & 3.03\\
        encoder-decoder         & 11.96             & 0.773 & 3.04\\
        \bottomrule
	\end{tabular}
	\vspace{\tabmargin}
\end{table}
}

\vspace{\secmargin}
\subsection{User study}
\begin{figure}[t]
\centering
\mfigure{1}{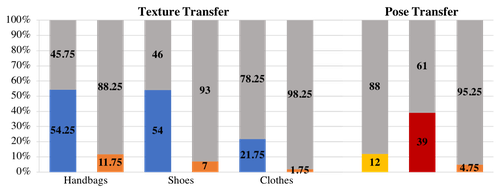} \\
\mfigure{0.9}{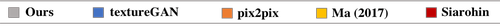} \\

\caption{\textbf{User Study.} The percentage of people that find our method more realistic respecting the input and guidance signal over state-of-the-art methods using pair-wise comparisons.}

\label{fig:user_study}
\end{figure}

We conduct a user study on pair-wise comparisons. We ask 100 subjects to answer 4 random pair-wise comparisons per task and dataset. We ask the subject to select the image that looks more realistic respecting the input and the given guidance signal. We show the user study results in \figref{user_study}.

\vspace{\secmargin}
\subsection{Limitation}
\begin{figure}[t]
\centering

\mfigure{0.15}{{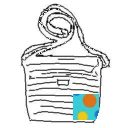}}\hfill
\mfigure{0.15}{{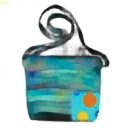}}\hfill
\mfigure{0.15}{{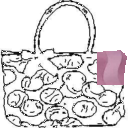}}\hfill
\mfigure{0.15}{{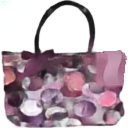}}\hfill
\mfigure{0.15}{{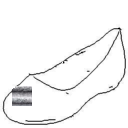}}\hfill
\mfigure{0.15}{{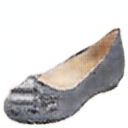}}\\

\vspace{\figcapmargin}
\captionof{figure}{\textbf{Failure examples.} When the guided patch does not match well with the given sketch, our model fails to hallucinate the given texture.
}
\label{fig:bag_failure}
\end{figure}

In the task of texture transfer, we observe a limitation of our work when the guidance patch does not go well with the input sketch. In such a case, the color of the guidance patch would propagate through the sketch without fully respecting its texture as shown in \figref{bag_failure}.

\section{Conclusion}
\label{sec:conclusions}

We have presented a new conditional scheme for guided image-to-image translation problems.
Our core technical contributions lie in the use of \emph{spatially varying} feature transformation and the design of \emph{bi-directional conditioning} scheme that allow the mutual modulation of the guidance and input network branches.
We validate the applicability of our method on various tasks.
While being application-agnostic, our approach achieves competitive performance with the state-of-the-art.
The generality of our method opens promising direction of incorporating a wide variety of constraints for image-to-image translation problems.

\noindent \textbf{Acknowledgment.} This work was supported in part by NSF under Grant No. 1755785. We thank the support of NVIDIA Corporation with the GPU donation.

{\small
\bibliographystyle{ieee_fullname}
\bibliography{main}
}

\end{document}